# Evaluation of Dataflow through layers of Deep Neural Networks in Classification and Regression Problems

Ahmad Kalhor[1]*, Mohsen Saffar[1], Melika Kheirieh[1], Somayyeh Hoseinipoor[1], Babak N. Araabi[1]

Emails: {akalhor, mohsen_saffar, mkheirieh, hoseinipoor0071,araabi}@ut.ac.ir

**Abstract** This paper introduces two straightforward and effective indices to evaluate the input data and the data flowing through layers of a feedforward deep neural network. For classification problems, the separation rate of target labels in the space of dataflow is explained as a key factor indicating the performance of designed layers in improving the generalization of the network. According to the explained concept, a shapeless distance-based evaluation index is proposed. Similarly, for regression problems, the smoothness rate of target outputs in the space of dataflow is explained as a key factor indicating the performance of designed layers in improving the generalization of the network. According to the explained smoothness concept, a shapeless distance-based smoothness index is proposed for regression problems. To consider more strictly concepts of separation and smoothness, their extended versions are introduced, and by interpreting a regression problem as a classification problem, it is shown that the separation and smoothness indices are related together. Through four case studies, the profits of using the introduced indices are shown. In the first case study, for classification and regression problems , the challenging of some known input datasets are compared respectively by the proposed separation and smoothness indices.  In the second case study, the quality of dataflow is evaluated through layers of two pre-trained VGG 16 networks in classification of Cifar10 and Cifar100.  In the third case study, it is shown that the correct classification rate and the separation index are almost equivalent through layers particularly while the serration index is increased. In the last case study, two multi-layer neural networks, which are designed for the prediction of Boston Housing price, are compared layer by layer by using the proposed smoothness index.

1. Introduction

To model classification and regression problems by neural networks for when there are challenging input data, it is required to design a sufficient number of layers for dimension reduction, disturbance filtering and providing invariances against distortions. By applying appropriate layers to challenging input data, an effective feature space is generated. Then, by adding one or two other fully connected layers, the required partitioning (or mapping) procedure on the feature space is performed and the classification (or regression) aim is satisfied. In the learning process of deep neural networks, one of the main concerns is how the data flowing through the layers should be evaluated in order to judge the performance of the designed layers.

Several concepts can be used to evaluate the neural network's performance. In classification problems, these concepts can interpret different aspects of misclassifications and correctness of a classifier. One general evaluation index that measures the performance of the network is loss functions as comparators between the real output and the estimated one. Regression losses such as mean square error, mean absolute error and mean bias error, as well as classification losses such as Hinge loss and cross entropy loss, are some common Loss functions in machine learning, [1], [2]. These loss functions are utilized to design network parameters using different methods such as back propagation approaches. However, by this means, one cannot measure the performance of each layer, individually. Even if the whole structure works fine and eligible, some hidden layers with negligible improvement could exist. The existence of such layers can lead to high redundancy, over parametrization, and low generalization. Therefore, there is Lack of a metric, which can evaluate not only the estimated labels, but also the input data, and the transient reformation of data after each layer.

[1]Authors are with the University of Tehran, College of Engineering, School of Electrical and Computer Engineering, Tehran, Iran
*Corresponding author

In [3], a method is established build on the information theoretic point of view that evaluate the hidden layers' performance and overcomes the flaws of cost function based methods. To this end, two aspects have been discussed in order to evaluate Deep Neural Network (DNN) based on the Information Bottleneck (IB) principle, [4]. DNN (Abbrev. should be used in the first usage and is used in all of the following cases) try to simplify input data passing through the layers by removing complexity and redundancy. This process improves the generalization of the system. DNN capture a minimal subset from input space, which is mostly related to the output. In fact, no structure can produce any new valuable information compared to the raw data so it should keep the relevant information as much as possible until the final layer.

When a DNN attempts to reach high simplicity in feature space, it might lose some important information available in the input about the output(confusing sentence). In other words, we may tolerate some complexity if we want to retain the maximum information. Accordingly, there exists a trade-off between those terms and a weighted summation covering both of the above facts can be regarded as a cost function. Tuning the weight factors can lead to different optimal spots in the feasible solutions regarding the designer's goal.

However, this method needs the joint distribution between input space and output, which is not computable in all circumstances, especially in lack of observations. Therefore, for some limited empirical data samples, this analysis may not result in a reliable conclusion. In addition, this method is a little professional to use by a normal user since the interpretation of weight factors are not explicit and adjusting the corresponding coefficients is not a straightforward task to do. In some parts of this paper, we introduce novel metrics, which hold the properties and benefits of the factors presented by the information bottleneck theory in a much simpler and appropriate way. There are some other works based on the information bottleneck. In [5], some solutions related to canonical correlation analysis based on Gaussian bottleneck have been given and in [6], some solutions for optimal predictive coding have been resulted by applying the Gaussian information bottleneck time series processes. An application of the bottleneck method to non-Gaussian sampled data by using the concept of past-future information bottleneck has been presented in [7]. In [8], the effect of the activation functions in compression phenomenon in deep neural networks has been studied. In [9], a rate-optimal estimator of mutual information has been proposed by which it is observed that the optimal hash-based estimator reveals the compression phenomenon in a wider range of networks with "ReLu" and "Maxpooling" activations.

Geometric metrics can also be employed in the neural network for design and evaluation. The fundamental idea behind them is the Euclidean distance between the points, which is very simple to follow. In [10] and [11], the authors use the K-means clustering method as a well-known geometric analysis in order to design the filter of a convolutional neural network. For this purpose, many centroids are extracted from the feature space using the k-means algorithm as a compressed representation of the raw data. Although this algorithm shows the promising result in a one-layer network, its performance for the deep neural network with multi hidden layers is not guaranteed due to the computational load of clustering methods. Apart from that, the performance of the method surely depends on the shape of the cluster and the method may not be applicable in different datasets.

In this paper, two novel geometric factors are proposed to prevail the aforementioned problems which are so convenient to use and can be applied to evaluate data, dataflow (representation of data after each hidden layer) and the result. The proposed geometric factors are based on two concepts of separation and smoothness respectively for classification and regression problems. After discussing their relation and defining their extensions, their profits are explained through some case studies. The organization of the paper is as follows: In Sections 2 and 3 evaluation of dataflow respectively in classification and regression problems are discussed. Some case studies, which shows the usage of the former indices, are presented in Section 4. Finally, in Section 5, concluding remarks are given.

## 2. Evaluation of dataflow in Classification problems

In a classification problem, it is aimed to build a classifier model, which can categorize input patterns (input data points) as a certain number of classes. In this paper, it is assumed that $n_C$ denotes the number of classes, which are represented by labels. The classifier model is a Feedforward deep neural network (DNN) with $n_{Layer}$ layers and each layer $L$ has $n_L$ neurons. Through a supervised learning process, the parameters of the DNN are learned in order that by applying Q input patterns $\{x^q\}_{q=1}^Q$ to the network, the resulted $\{\hat{l}^q\}_{q=1}^Q$ have high equivalency to the target labels $\{l^q\}_{q=1}^Q$, where $l^q \epsilon \{1,2,...,n_C\}$. A classification process in a DNN can be explained in two parts: in the first part ("feature extracting" part), the nullities, disturbances, and distortions of the input patterns are filtered and a new space including effective features, indicated by **z**, is appeared. In the second part ("partitioning" part) by making boundaries around the local regions including patterns with the same labels, i.e. partitioning in the feature space **z**, the label of each input pattern is revealed at the output of the network (Figure 1).

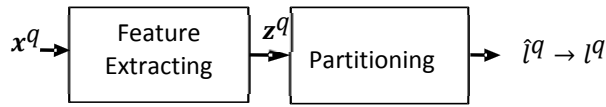

**Figure 1:** A diagram of DNN with two parts for classification purpose

### 2.1 Separation concept

In the classification of many real-world datasets, due to disturbances and distortions in different scales, the intake patterns are in casual positions in the vector space. This leads patterns with different labels become adjacent together and as a result, the regions of patterns with the same labels cannot form anymore.

However, to provide generalization in a classification problem, it is required that all patterns with the same labels become near together or equally patterns with different labels become separated from each other. Actually, in such condition, the required boundaries or the number of neurons at "partitioning" part becomes low and as a result, the sensitivity of the network to uncertainties in the feature space is reduced and eventually the network with a lower number of parameters can provide better generalization. In contrast, if patterns with different labels become near together in the feature space, the required boundaries, as well as the number of neurons in "partitioning" part, will significantly increase and consequently, over parametrization is resulted. In such state by adding a limited uncertainty to an incoming pattern, its assigned true label may change to a false one. This means that the classifier has high sensitivity to the uncertainties in the feature space and the generalization of the network lose (??). To explain the concept of "separation" between patterns with different labels, a simple hand-made illustrative example is shown in Figure 2. To show details both input patterns and feature points are shown in a two-dimensional space.

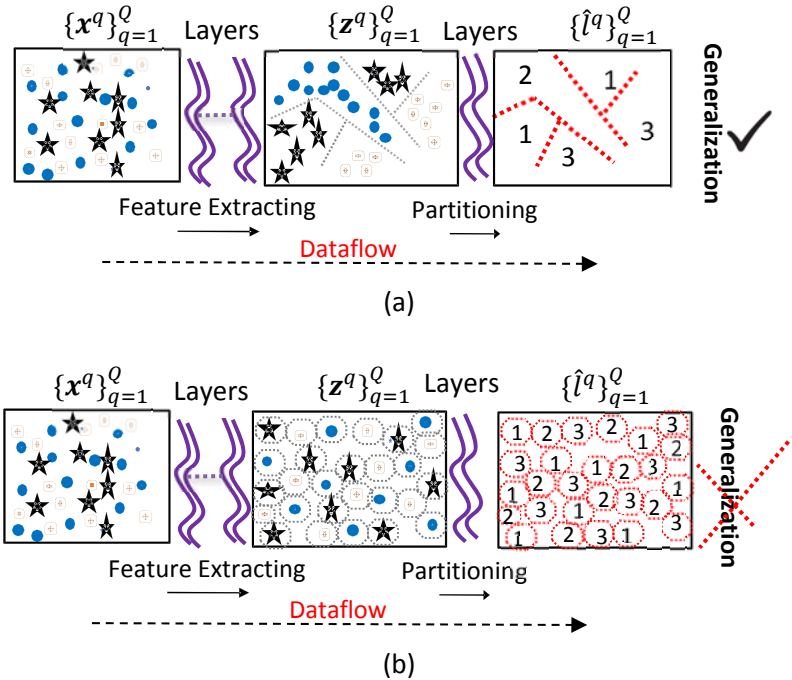

**Figure 2:** An illustrative handmade example to explain the concept of separation in getting generalization in a classification problem

As it is seen in Figure 2 (a), at the entrance of the network, the intake patterns with different labels (indicated by stars, bullets, and squares) are in causal positions. However, by designing an appropriate "feature extracting" part including a sufficient number of layers, the feature points with the same labels come nearer together within some local regions in the feature space. In contrast, in Figure 2 (b), since patterns with the same labels are still far from each other at feature space, the required boundaries for "partitioning" as well as the required number of neurons increases and due to over-parametrization, the generalization becomes low. From Figure 2(b), it is understood that by a little uncertainty in getting feature point of an incoming pattern, its true label may change abruptly, but such sensitivity is not understood in the example Figure 2(a).

In DNNs, due to using continues activation functions and a frequent number of batch normalization layers, following assumptions are considered: (1) each layer causes continuous changes to dataflow and (2) the scales of dataflow at different dimensions are almost identical. According to these assumptions and with respect to aforementioned necessity to get high generalization, it is expected that the separation rate between feature points with non-equal labels is increased layer by layer in order to get high separation rate at end of "feature extracting" part. Such concept of separation rate can be computed as a numerical index to evaluate the input data and dataflow at each layer of the deep neural network. In the following section, an easy shapeless distance-based separation index is proposed.

## 2.2 Separation Index

Here, in a classification problem, where $\{x^q, l^q\}_{q=1}^{Q}$ denotes the pairs of input patterns and output target labels, and $\{x_L^q\}_{q=1}^{Q}$ represents $\{x^q\}_{q=1}^{Q}$ as dataflow in layer $L \in \{1, .., n_{Layer}\}$, the separation index (SI), at layer $L$, is defined as follows:

(1)
$$\mathrm{SI}(\{x_L^q\}_{q=1}^{Q}, \{l^q\}_{q=1}^{Q}) = \frac{1}{Q}\sum_{q=1}^{Q} \varphi\left(l^q - l^{q_{near}^L}\right) \quad \varphi(v) = \begin{cases} 1 & v = 0 \\ 0 & v \neq 0 \end{cases}$$
$$q_{near}^L = \arg\min_{h}\|x_L^q - x_L^h\| \quad h \in \{1,2,..,Q\} \text{ and } h \neq q$$

where $\varphi(\cdot)$ denotes Kronecker delta function. From (1), it is understood that the SI is defined as a normalized number of neighboring pairs of patterns with equal labels at layer $L$. Since the maximum number of neighboring pairs with the same labels is Q, it is concluded that $0 \leq \mathrm{SI}(\{x_L^q\}_{q=1}^{Q}, \{l^q\}_{q=1}^{Q}) \leq 1$. It is feasible to say that for when $\mathrm{SI}(\{x_1^q\}_{q=1}^{Q}, \{l^q\}_{q=1}^{Q})$ is nearer to "one", the separation between patterns with different classes initially exists and a DNN should be designed mainly for "partitioning" part. In contrast, when $\mathrm{SI}(\{x_1^q\}_{q=1}^{Q}, \{l^q\}_{q=1}^{Q})$ is nearer to "zero", the input data is challenging and the designer should use a sufficient number of layers to prepare more effective features with higher SI.

As it was explained, in an appropriately designed DNN, it is expected that the smoothness index (SmI) increase almost uniformly through layers of "feature extracting" part with different trends (Figure 3). In Figure 3, $\mathrm{SI}_0 \geq 0$ and $\mathrm{SI}_{max} \leq 1$ respectively indicate the initial and supremum separating index of dataflow. In this figure, there are three increasing trends with different, initial slopes, rise times and settling times.

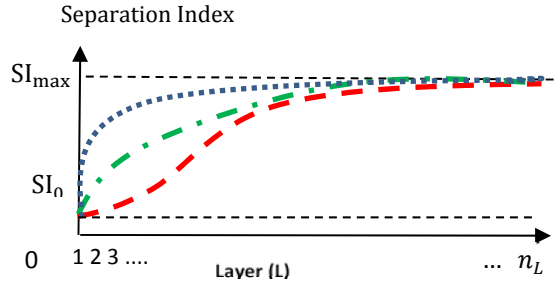

**Figure 3:** Some expected increasing trends of the separated index through the layers in an appropriately designed MLNN.

## 2.3. Strictly Separation Index

In order to evaluate the stricter concept of separation, by considering labels of $r$ ($r > 1$) nearest feature points (of dataflow in layer $L$), the former introduced SI in (2) is generalized as follows:

(2)
$$\mathrm{SI}_r(\{x_L^q\}_{q=1}^{Q}, \{y^q\}_{q=1}^{Q}) = \frac{1}{Q}\sum_{q=1}^{Q}\prod_{k=1}^{r} \varphi\left(l^q - l^{q_{near}^L(r)}\right)$$
$$[q_{near}^L(1) \; q_{near}^L(2) \ldots q_{near}^L(Q-1)] = \arg\mathrm{sort}_{h, h\neq q \text{ ascending}} \|x_L^q - x_L^h\|$$

where $\varphi(\cdot)$ denotes Kronecker delta function. From (2), it is understood that the strictly SI with order $r = 1$ is the same SI introduced in (1) and for order $r > 1$ the strictly SI is less than or equal to the SI with order $r = 1$. In fact, by increasing the order $r$ we can evaluate more strictly SI. In the case, the strictly SI remains high for when order $r$ increases, a greater number of feature points with the same labels are localized in a region and the partitioning becomes easier.

## 3. Evaluation of Dataflow in Regression problems

In a regression problem, it is aimed to build a model, which maps input patterns to the target outputs. Assuming the regression model is a feedforward DNN, the parameters of the network are learned in order that by applying $\{x^q\}_{q=1}^{Q}$ to the network, the resulted outputs $\{\hat{y}^q\}_{q=1}^{Q}$ have high equivalency with target outputs $\{y^q\}_{q=1}^{Q}$. It is assumed that the dimension of the output is $m \geq 1$ and $y^q = [y_1^q, y_2^q, \ldots, y_m^q]$. In addition, it is assumed *each* output $y_j$ ($j = 1,2,\ldots,m$) has been whitened, individually, where the mean and the variance of its samples, $\{y_j^q\}_{q=1}^{Q}$ are "zero" and "one", respectively.

In this section, the concept of smoothness is explained and then after a SmI is introduced for regression problems. By extending the SmI, strictly SmI is defined and finally, the relation between SI and SmI is explained.

### 3.1 Smoothness concept

Here, similar to a classification problem, it is desired that a deep feedforward neural network is learned in two parts. At the first part ("feature extracting" part), the input space ($x$) is transformed into effective feature space ($z$) in which nullities and disturbances have been removed. In the second part ("mapping" part), it is desired that based on the feature space $z$, a regression function is approximated in which by applying $\{z^q\}_{q=1}^{Q}$, the resulting outputs, $\{\hat{y}^q\}_{q=1}^{Q}$, make high equivalency with targets outputs, $\{y^q\}_{q=1}^{Q}$ (Figure 4).

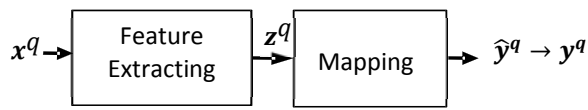

**Figure 4:** A diagram of a neural network for regression purpose

In a regression problem, smoothness of the approximated function on the feature space is very important, because it avoids erratic variations of the output by little changes of feature points at feature face $z$. A necessary condition for smoothness is that for each pair of effective feature points: $z^{q_1}$ and $z^{q_2}$ when they become very near together $\|z^{q_1} - z^{q_2}\| \to 0$, the corresponding outputs $y^{q_1}$ and $y^{q_2}$ become very near together, $\|y^{q_1} - y^{q_2}\| \to 0$. Such condition allows minimizing the sensitivity of output against input variations and consequently the regression function get a good generalization. In Figure 5, it is tried to explain the importance of the smoothness concept through an illustrative example. As it is seen in Figure 5 (a), from initial disturbed input space, by designing an appropriate "feature extracting" part a feature space results, on which a regression function with high enough smoothness is built. In such an example, it is expected that for any incoming new data point, the nearest neighbors make outputs with low error with the output of the considered incoming data point. In contrast, in Figure 5 (b), a feature space is appeared, on which an unsmooth jaggy function has been fitted. It is expected that the generalization of such function becomes low because the output is very sensitive to little variations on feature space.

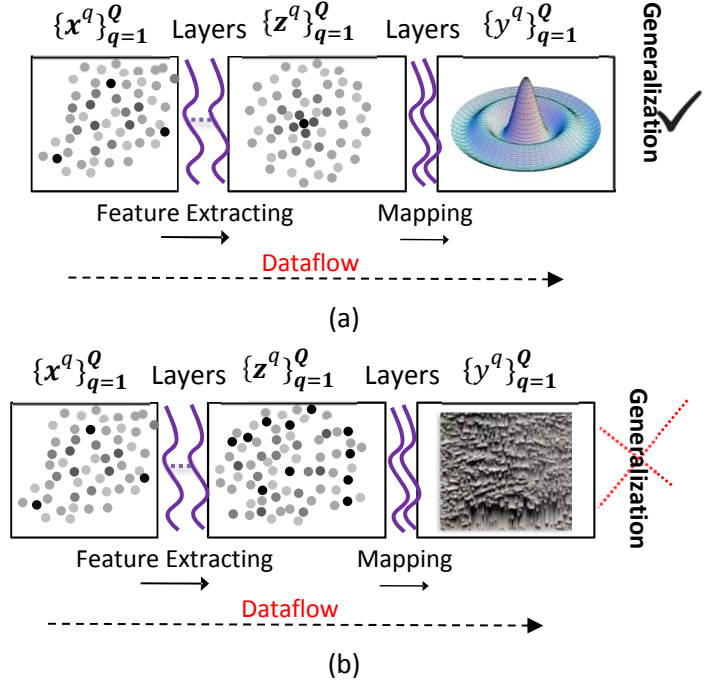

**Figure 5:** An illustrative handmade example to explain (a) the effect of smoothness in getting generalization and (b) lack of smoothness in missing generalization, in a regression problem

If it is assumed that each layer, due to using continues activation functions, has continued functionality on dataflow, the smoothness should increase gradually through layers of the network, in order to get high enough smoothness at end of dataflow. To evaluate the smoothness of the input data and dataflow at each layer of the DNN, a straightforward numerical index is proposed in the following section.

**3.2 Smoothness index**

Here, for a regression problem, where $\{x^q, y^q\}_{q=1}^Q$ denotes the pairs of input patterns and real outputs, the SmI at layer $L \in \{1, .., n_{Layer}\}$ is defined as follows:

(3)
$$\text{SmI}(\{x_L^q\}_{q=1}^Q, \{y^q\}_{q=1}^Q) = \frac{1}{Q} \sum_{q=1}^Q \exp\left(-\beta\left(d_{q_{near}^L}^y - d_{q_{min}}^y\right)\right) \quad \beta > 0$$

$$d_{q_{near}^L}^y = \left\|y^q - y^{q_{near}^L}\right\|^2 \qquad d_{q_{min}}^y = \|y^q - y^{q_{min}}\|^2$$

$$q_{near}^L = \arg\min_{h, h \neq q} \|x_L^q - x_L^h\| \qquad q_{min} = \arg\min_{h, h \neq q} \|y^q - y^h\| \qquad h \in \{1,2,..,Q\}$$

where $\beta$ denotes the smoothness coefficient which determines the rate of smoothness variation when distances of neighboring points changes. Here in this paper, in order to assign very low SmI (about 0.01 or 0.02) when $\left(d_{q_{near}^L}^y - d_{q_{min}}^y\right)$ is about the variance of the output (variance of each output is assumed "one"), $\beta = 4$ has opted.

From (4), it is understood that $\text{SmI}(\{x_L^q\}_{q=1}^Q, \{y^q\}_{q=1}^Q)$ is the mean of output distances of all nearest neighboring pairs of dataflow at layer $L$. Since the arguments of all exponential terms are not positive, it is seen that $0 \leq \text{SmI}_A \leq 1$. It is expected that for well-behaved input $\text{SmI}_A$ increases to "one" and for challenging data when there are disturbances, distortions, and nullities, $\text{SmI}_A$ deceases to "zero". As a result, for when $SmI_A$ is near to "one", the input data makes the smooth output and only some fully connected layers should be designed for mapping purpose. In contrast, when $SmI_A$ is near to "zero", the input data is challenging and the designer should use a sufficient number of layers to prepare more effective features with higher SmI. One can examine that in an appropriately designed network whose "feature extracting" part has $n_L$ layers, the $SmI_A$ increases gradually through layers (Figure 6). In Figure 6, $SmI_0$, and $SmI_{max} \leq 1$ indicate respectively initial and final SmI of the dataflow through layers of the deep neural network. In addition, there are three increasing trends with different initial slopes, rise times and settling times.

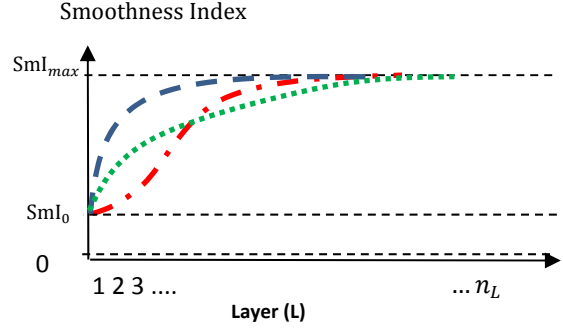

**Figure 6:** Some expected increasing trends of smoothness index through layers in an appropriately designed network

### 3.3 Strictly Smoothness index

The former introduced SmI uses only one nearest neighbor of each certain data point. In order to provide more strictly concept of smoothness for evaluation of dataflow in a regression function, we can consider $r > 1$ nearest neighbors of each certain data point. For this mean, the following index is proposed:

$$
(4) \quad SmI_r(\{x_L^q\}_{q=1}^Q, \{y^q\}_{q=1}^Q) = \frac{1}{Q} \sum_{q=1}^Q \prod_{k=1}^r \exp\left(-\beta d_{q_{near}^L(k)}^x \left(d_{q_{near}^L(k)}^y - d_{q_{min}(k)}^y\right)\right)
$$

$$
[q_{near}^L(1) \; q_{near}^L(2) \ldots q_{near}^L(Q-1)] = \underset{h, h \neq q}{\arg} \underset{ascending}{sort} \|x_L^q - x_L^h\|, \quad [q_{min}(1) \; q_{min}(2) \ldots q_{min}(Q-1)]
$$

$$
= \underset{h, h \neq q}{\arg} \underset{ascending}{sort} \|y^q - y^h\|
$$

where r denotes the order of smoothness, $d_{q_{near}^L(k)}^x \leq 1$ and $d_{q_{near}^L(k)}^y$ are defined as follow:

$$
(5) \quad d_{q_{near}^L(k)}^x = \begin{cases} \dfrac{\left\|x_L^q - x_L^{q_{near}^L(1)}\right\|^2}{\left\|x_L^q - x_L^{q_{near}^L(k)}\right\|^2} & if \; \left\|x_L^q - x_L^{q_{near}^L(k)}\right\| > 0 \\ 1 & otherwise \end{cases}, \; d_{q_{near}^L(k)}^y = \left\|y^q - y^{q_{near}^L(k)}\right\|^2
$$

For $r = 1$, the former introduced SmI in (4) results. For $1 < r < Q$ in order to evaluate (more strictly) smoothness of the dataflow, it is required that for each example $(x_L^q, y^q)$, $r$ outputs corresponded to $r$ nearest neighbors to $x_L^q$, have lower distances to $y^q$. Assuming $\mathrm{SmI}_A^L(0) = 1$, it can be shown easily that for $= 1,2 \ldots Q-1$, $\mathrm{SmI}_r(\{x_L^q\}_{q=1}^Q, \{y^q\}_{q=1}^Q) \leq \mathrm{SmI}_{r-1}(\{x_L^q\}_{q=1}^Q, \{y^q\}_{q=1}^Q) \leq 1$ and for $a \neq 0$, $\mathrm{SmI}_r(\{ay^q + b\}_{q=1}^Q, \{y^q\}_{q=1}^Q) = 1$. This means that if in an ideal state, the dataflow points at layer $L$ become an arbitral scaled and shifted of the target labels, the SmI of dataflow is maximum.

### 3.4 Relation between separation index and smoothness index

Here, by imposing some constraints, a regression problem is represented as a classification problem and it is shown that the SI and SmI are related together. Actually, by quantizing the output of a regression problem and considering that each quantized level represents an individual class and there are (1) perfect nearness between two patterns with the same class and (2) no nearness between two patterns with different classes, a regression problem can be represented as a classification problem. For the sake of simplicity, it is assumed that the output of a regression problem is scalar. Considering $y_{min}$ and $y_{sup}$ respectively denote the minimum and the supremum of the output, the output can be quantized to $n_C$ sequenced levels with equal distance $\rho = (y_{max} - y_{min})/n_C$. Consequently, each output $y^q$ can be indicated by label $l^q \in \{1,2,\ldots,n_C\}$, which is computed as follows:

(6) $\qquad l^q = label(y^q) \quad$ when $\quad y_{min} + (l^q - 1)\rho \leq y^q < y_{min} + l^q$

Figure 7 shows a simple example where the real output signal (dot line) has been quantized into $n_c = 5$ levels (solid line). From Figure 7 it is understood that the classification problem converges to the former regression problem if $n_c \to \infty$.

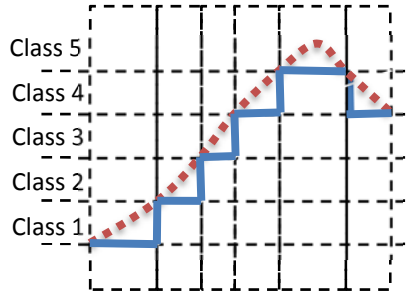

**Figure 7:** A quantized regression problem by $n_c = 5$ levels

If we assume that, there is no meaningful nearness between classes, regarding to (3) for each $(x_L^q, l^q)$, its distance to a member of the same class is zero and its distance to a member of other classes is infinity. Assuming each class has at least two members then $d_{q_{min}}^y = 0$. Hence, equation (2) can be rewritten as follows:

(7) $\qquad \mathrm{SmI}(\{x_L^q\}_{q=1}^Q, \{l^q\}_{q=1}^Q) = \dfrac{1}{Q}\sum_{q=1}^{Q} \exp(-\beta d_{q_{near}^L}^y) \quad d_{q_{near}^L}^y = \begin{cases} 0 & \text{if } l^q = l^{q_{near}^L} \\ \infty & \text{otherwise} \end{cases}$

Comparing (7) and (1), it is revealed that the new derived SmI for the represented classification problem and the former defined SI are equal together. The above-mentioned relation between SI and SmI can be generalized easily for both strict versions of separation and smoothness indices with equal orders.

## 4. Case Studies

In this section, some applications of using former introduced separation and smoothness indices are shown through some case studies. In the first case study, the challenging of some known data sets, which are used in classification and regression problems, are evaluated and compared. In the second case study, the quality of dataflow as well as the performance of different layers of two convolutional neural networks, which are designed for classification of "Cifar10" and "Cifar100", are evaluated. In the third case study, the correlation between SI and correct classification rate through convolutional layers, are studied and in the last case-study, Two multi-layers neural; networks, which are designed for the prediction of Boston housing price, are evaluated and compared.

### 4.1 Data Evaluation in classification and Regression problems

There are some known datasets usually used as benchmarks in classification and regression problems. Here, a few low dimensional datasets are chosen to be evaluated by the proposed evaluation indices. Table 1 presents the results of applying the SI to data sets: "MNIST-Digits", "MNIST- Fashion", "Cifar 100" and "Cifar 10" for 5000 training data points.

Table1: Evaluation of some known classification datasets by using the separation index

| Dataset | Number of Classes | Separation Index |
|---|---|---|
| MNIST Digits | 10 | 0.97372 |
| MNIST Fashion | 10 | 0.85423 |
| Cifar-10 | 10 | 0.2636 |
| Cifar-100 | 100 | 0.17446 |

From the computed SI in Table 1, one can order the data sets from more to less challenging (left to right) as it is seen in Figure 8.

"Cifar 100" > "Cifar 10" > "MNIST − Fasion" > "MNIST − Digits"

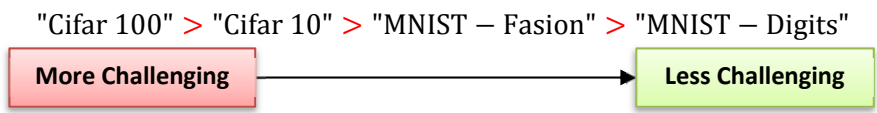

**Figure 8:** The ordered classification data sets from more challenging to less challenging

Table 2 presents the results of applying the SmI with orders $r = 1,2,3$ to data sets: "Gold price", "Oil Price", "electrical load" and "daily and monthly sunspot number". For each time series, in order to predict the time series at the current time, $y_t$, a certain number of former lags are considered as input $x_t = [y_{t-1}, y_{t-2}, \ldots y_{t-n}]$.

From the presented smoothness indices in Table 2, one can order the data sets from more to less challenging (left to right) as it is seen in Figure 9.

"Daily Sunspot number " > "Monthly Sunspot number " > "Electrical loads" > "Gold price" > "Oil Price"

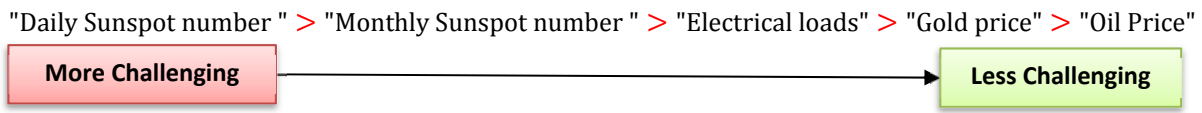

**Figure 9:** The ordered time series data sets from more challenging to less challenging

In Table 2, it is seen that the SmI of each time series is decreased when the smoothness order is increased.

Table 2: Evaluation of some benchmark time series by using three first orders of smoothness index

| Time Series | Interval | Input $x_t$ | Smoothness Index order | | |
|---|---|---|---|---|---|
| | | Number of lags | $r=1$ | $r=2$ | $r=3$ |
| Gold price(Monthly-mean) | 1950-1918 [12] | n=12 (year cycle) | 0.8392 | 0.7829 | 0.7089 |
| Oil Price(Weekends) | 2005- 2019 [13] | n=13 (season cycle) | 0.8757 | 0.8090 | 0.7245 |
| Electrical load s(daily-maximum) | 1997-1998 [14] | n=7 (week cycle) | 0.7093 | 0.5527 | 0.4458 |
| Sunspot number (Daily-total) | 2016-2018 [15] | n=7 (week cycle) | 0.5950 | 0.4293 | 0.3312 |
| sunspot number(Monthly-mean) | 1950-1918 [15] | n=12 (year cycle) | 0.6563 | 0.5037 | 0.3960 |

**4.2 Dataflow Evaluation in the classification of Cifar10 and Cifar100 by "VGG 16" Network[2]**

Data sets "Cifar10" and "Cifar100", due to their low dimensions and their challenges, have been used frequently as benchmarks in classification approaches. The convolutional neural network "**VGG** 16" ([16]), which has been utilized successfully in the classification of "Cifar10" and "Cifar100", [17]. Here, it is aimed to evaluate dataflow through layers of two pre-trained "VGG 16" networks (Figure 10).

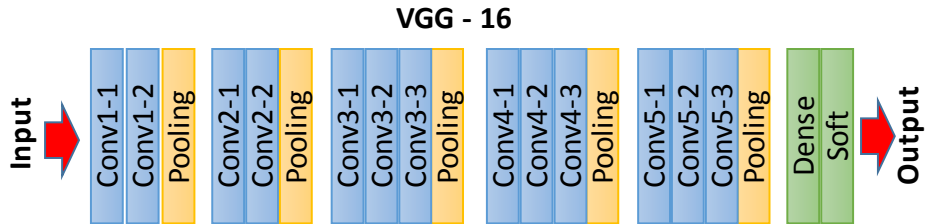

**Figure 10:** the architecture of "VGG 16" network

Since the evaluation of whole training and test dataflow through all layers is time-consuming, so it is decided to study the sensitivity of the index against the number of data points. Figure 11 shows the plot of the computed SI versus the different number of shuffled data points in both "Cifar10" and "Cifar100". As it is seen, the SI almost increases uniformly, particularly when the number of training data points increases. However, for a very low number of data points, the index may have non-smooth variations.

With regarding to Figure 11, in order to have more reliable results, the dataflow is evaluated respectively by 20000 and 30000 shuffled data points of training data sets for "Cifar10" and "Cifar100"; however, for test data points of "Cifar10" and "Cifar100", all available 10000 data points have been applied in evaluation.

Tables 3 presents the details of layers of "VGG 16" network in the classification of "Cifar10" and "Cifar100".

---

[2] The source *code* can be accessed from*:* https://github.com/melika-kheirieh/Seprability-index-CNN

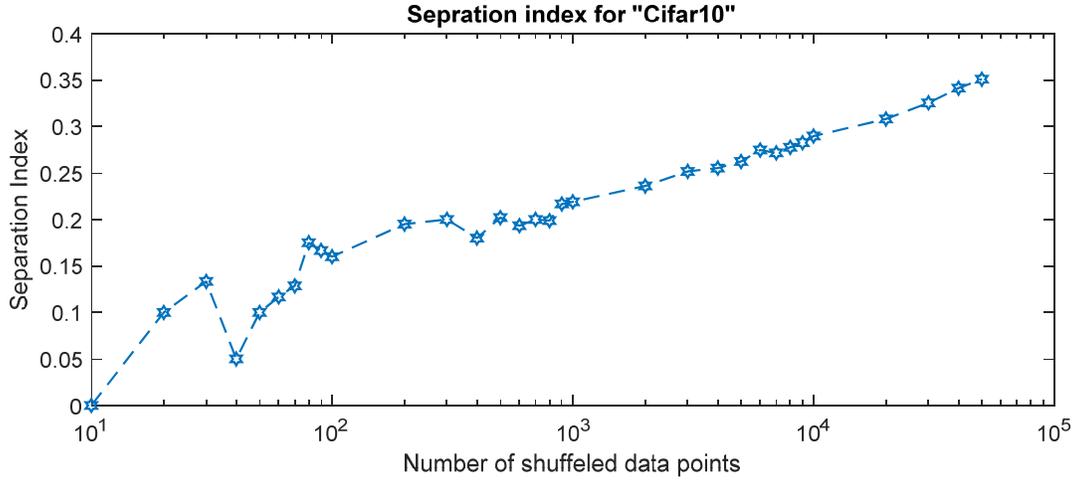

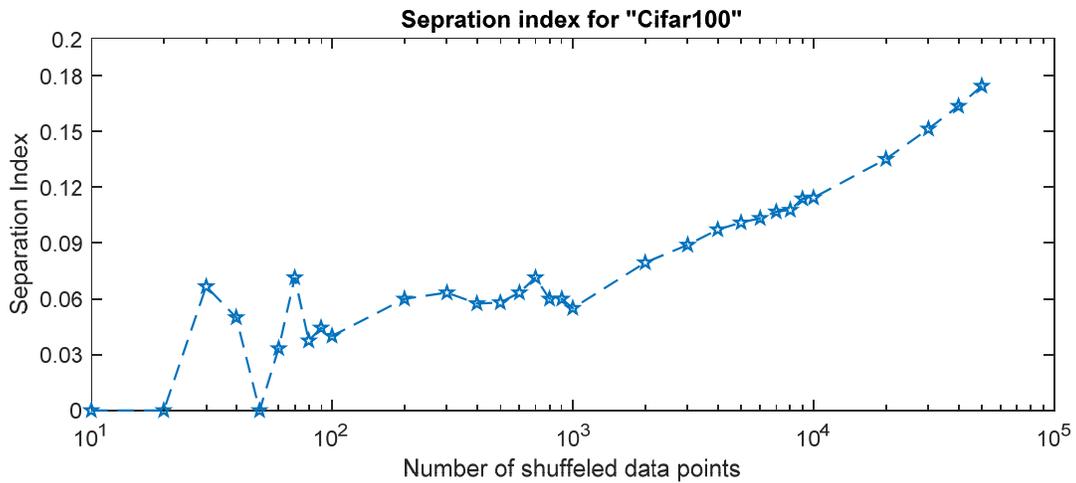

**Figure 11:** the plot of separation index versus the different number of shuffled data points in both "cifar10" (a) and "Cifar100" (b)

Table 3: The whole layers of "VGG 16" network in the classification of "Cifar10" and "Cifar10"

| Layer Number | Layer Name | Layer Number | Layer Name | Layer Number | Layer Name | Layer Number | Layer Name |
|---|---|---|---|---|---|---|---|
| 1 | **Conv2D** | 16 | MaxPooling | 31 | Batch Nor. | 46 | Activation |
| 2 | Activation | 17 | **Conv2D** | 32 | Dropout | 47 | Batch Nor. |
| 3 | Batch Nor. | 18 | Activation | 33 | **Conv2D** | 48 | **Dropout** |
| 4 | Dropout | 19 | Batch Nor. | 34 | Activation | 49 | **Conv2D** |
| 5 | **Conv2D** | 20 | Dropout | 35 | Batch Nor. | 50 | Activation |
| 6 | Activation | 21 | **Conv2D** | 36 | Dropout | 51 | Batch Nor. |
| 7 | Batch Nor. | 22 | Activation | 37 | **Conv2D** | 52 | *MaxPooling* |
| 8 | *MaxPooling* | 23 | Batch Nor. | 38 | Activation | 53 | **Dropout** |
| 9 | **Conv2D** | 24 | Dropout | 39 | Batch Nor. | 54 | Flatten |
| 10 | Activation | 25 | **Conv2D** | 40 | *MaxPooling* | 55 | Dense |
| 11 | Batch Nor. | 26 | Activation | 41 | **Conv2D** | 56 | Activation |
| 12 | Dropout | 27 | Batch Nor. | 42 | Activation | 57 | Batch Nor. |
| 13 | **Conv2D** | 28 | *MaxPooling* | 43 | Batch Nor. | 58 | **Dropout** |
| 14 | Activation | 29 | **Conv2D** | 44 | Dropout | 59 | Dense |
| 15 | Batch Nor. | 30 | Activation | 45 | **Conv2D** | 60 | Activation |

Figures 12 and 13 show the plots of computed SI (both training and test data) for layers of the network respectively used in the classification of "Cifar10" and "Cifar100".

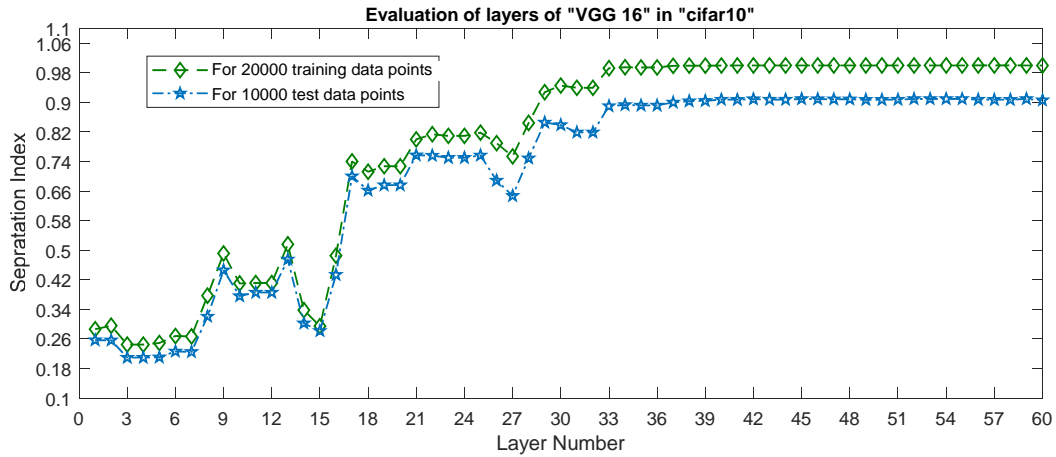

**Figure 12:** Evaluation of dataflow through layers of "VGG 16" network in the classification of Cifar10

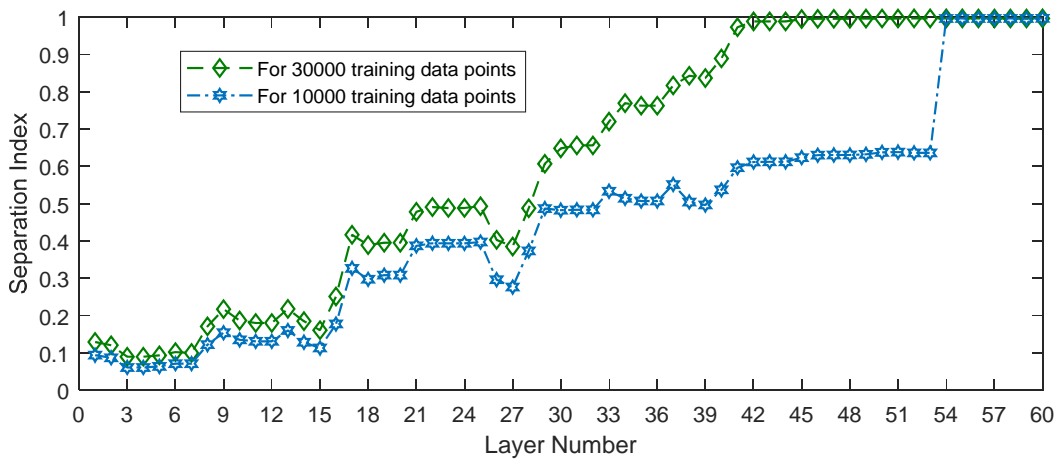

**Figure 13:** Evaluation of dataflow through layers of "VGG 16" network in the classification of Cifar100

Figures 12 and 13 reveals some details about the training or test dataflow at different layers of two designed "VGG 16" networks:

(1) Ignoring some undershoots on SI, one can say that the SI increases almost uniformly through the layers.
(2) SI plot for training detest is upper than SI plot for the test dataset.
(3) The SI plot meets its maximum value, some layers before the end of dataflow but for the test data, it occurs at further layers.
(4) "Maxpooling" layers are important layers: just after almost all "Maxpooling" layers there are fast increasing trends on SI.
(5) For the cases, there are two or three-sequenced convolution layers, in the first convolution layer, the SI increases fast but in the last one, there may be little undershooting on the index.

(6) The SI in the classification of the "Cifar100" (in comparison to "Cifar10") has less smoothness particularly about the SI of test dataflow.

**4.3 Relation Between Separation Index and Correct Classification Rate through layers**

In this section, the relation between the SI and correct classification rate (CCR), in the classification of "Cifar10", is studied. For this mean, after each convolution layer of a pertained "vgg-16", two dense layers are added in order to predict the true labels. To train the dense layers, 10000 data points of training dataset are used and 10000 data points of test dataset are considered for validation purpose. After two dense layers, a batch-normalization layer is utilized and after them, a softmax layer is applied. Considering sum of squared error as loss function, "Adam" optimizer with more than 100 epochs has been utilized. Table 4 presents the details of the dense layers and number of epochs at each convolution layer.

Table 5 presents the SI and CCR at each convolution layer for both training and test datasets and Figure 14 shows the plots of separation indices and CRR along with convolutional layers.

As it is understood from Figure 14, SI and CRR are almost equivalent particularly when SI increases at both training and test data sets. According to this result, SI even in non-strictly version is suitable to evaluate the correct classification rate.

Table 4: The designed dense layers including the number of neurons and number of epochs, in order to compute the correct classification rate after each convolution layer

| Convolution Layer | First Dense Layer | Second Dense Layer | Epochs |
|---|---|---|---|
| | Number of neurons | Number of neurons | |
| 1,2 | 280 | 120 | 170 |
| 3,4,5,6,7 | 250 | 100 | 170 |
| 8,9 | 220 | 50 | 130 |
| 10 | 200 | 80 | 130 |
| 11 | 150 | 70 | 100 |
| 12,13 | 100 | 50 | 100 |

Table 5: The separation index and correct classification rate at each convolution layer for both training and test datasets

| Convolution Layer | SI train | SI test | Train CCR | Test CCR |
|---|---|---|---|---|
| 1 | 0.216 | 0.2561 | 10.32 | 10 |
| 2 | 0.207 | 0.2261 | 46.47 | 38.9 |
| 3 | 0.3575 | 0.3760 | 67.55 | 58.35 |
| 4 | 0.3915 | 0.3022 | 59.65 | 53.94 |
| 5 | 0.6155 | 0.6617 | 78.79 | 70.99 |
| 6 | 0.7085 | 0.7559 | 87 | 79 |
| 7 | 0.7175 | 0.6884 | 76.25 | 69.86 |
| 8 | 0.8845 | 0.8386 | 93.77 | 86.23 |
| 9 | 0.9835 | 0.8918 | 99.19 | 91.59 |
| 10 | 0.996 | 0.9041 | 99.03 | 90.88 |
| 11 | 1 | 0.9074 | 99.92 | 92.37 |
| 12 | 1 | 0.9092 | 99.87 | 92.4 |
| 13 | 1 | 0.9070 | 99.95 | 92.88 |

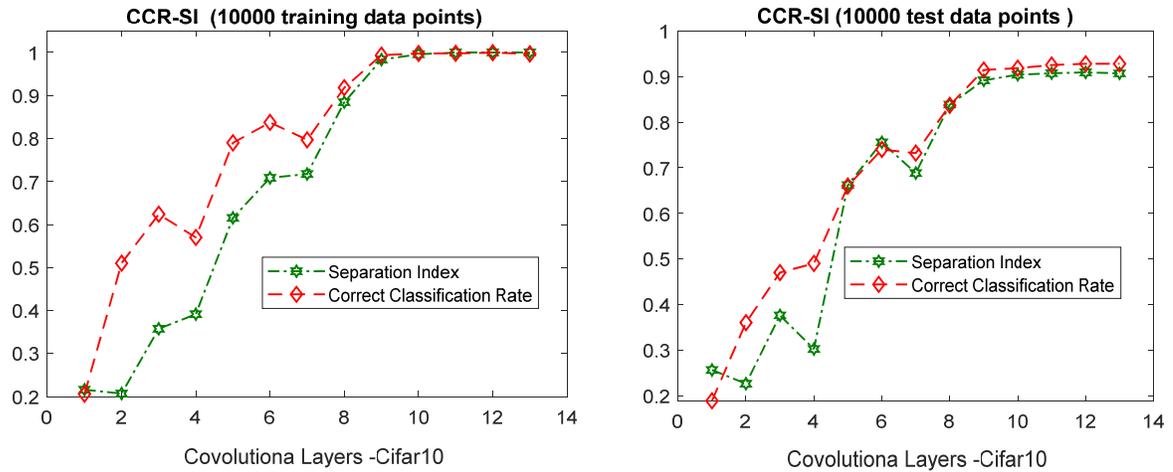

**Figure 14:** the plots of separation indices and correct classification rates at convolution layers in a pertained "vgg-16" network utilized for classification of "Cifar10".

### 4.4 Dataflow Evaluation in the prediction of Boston housing price

In this case study, evaluation of dataflow is considered through layers of two designed neural networks in a regression problem. For this mean prediction of Boston housing price is aimed by using multi-layer neural networks (MLNNs). The dataset of Boston housing price contains information collected by the U.S Census Service concerning housing in the area of Boston Mass [18]. At the available training (test) data, there are 404 (102) $n = 13$ dimensional feature points to predict the Boston housing price. To design MLNNs, after data normalization, two hidden layers has been trained unsupervised in order to dimension reduction.

Then after encoding layers and using the resulted feature space, at the first strategy (MLNN1), a linear model is optimized by the least square technique in order to minimize the sum of squared errors. At the second strategy (MLNN2), two fully connected layers including 100 and 50 neurons are trained by stochastic point based error back propagation method (with learning rate 0.02 and 2000 epochs) in order to minimize sum of squared error. After training MLNN1 and MLNN2, the validation of two networks are examined by test data. Table 6 represents the computed root mean squared error (RMSE) of both networks in training and test data. As it is seen the MLNN2 is a far better model to predict Boston housing price.

Table 6: the computed RMSE for two designed MLNN1 and MLNN2 for both training and test data

| Network | RMSE | |
|---|---|---|
| | Training data | Test data |
| MLNN1 | 0.4061 | 0.4015 |
| MLNN2 | 0.0478 | 0.0986 |

Table 7: Evaluation and comparison of dataflow through layers of two MLNNs in the prediction of Boston housing price

| Network | Topology | Normalization Layer | | Encoding Layers | | Mapping Layers | | |
|---|---|---|---|---|---|---|---|---|
| | Layers: | Layer1 (input) | Layer2 | Layer3 | Layer4 | Layer5 | Layer6 | Layer7 |
| MLNN1 | Neurons: | 13 | 13 | 11 | 8 | 1 | - | - |
| | Activation: | - | - | sigmoid | sigmoid | linear | | |
| | **SmI (train)** | 0.5994 | 0.7005 | 0.7119 | 0.7399 | 0.6748 | - | - |
| | **SmI (test)** | 0.5519 | 0.6049 | 0.5899 | 0.6022 | 0.4786 | - | - |
| MLNN2 | Neurons: | 13 | 13 | 11 | 8 | 100 | 50 | 1 |
| | Activation: | - | - | sigmoid | sigmoid | sigmoid | sigmoid | sigmoid |
| | **SmI (train)** | 0.5994 | 0.7005 | 0.7119 | 0.7399 | 0.7476 | 0.7467 | 0.8064 |
| | **SmI (test)** | 0.5519 | 0.6049 | 0.5899 | 0.6022 | 0.6202 | 0.6382 | 0.6668 |

However, if someone likes to evaluate that each layer of the designed networks how well improved the dataflow, the former introduced SmI can be used. Table 8 presents the summarized results of applying SmI with order $r = 1$ to dataflow through layers MLNN1 and MLNN2.

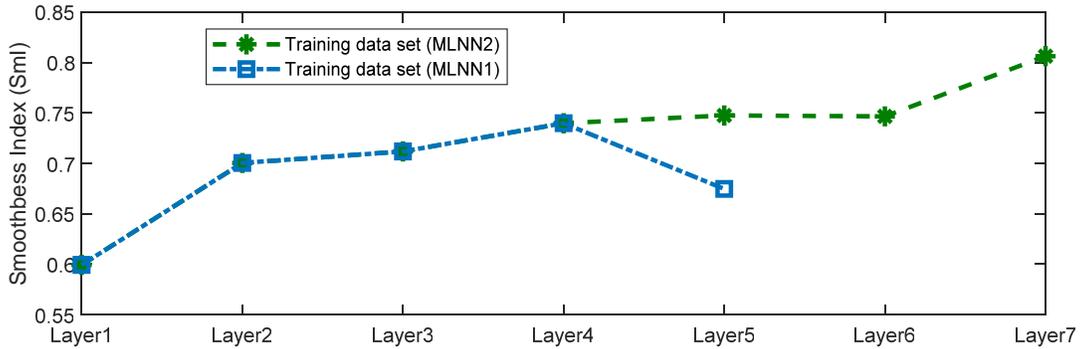

**Figure 15:** Evaluation of training dataflow through layers of two MLNNs in the prediction of Boston housing price.

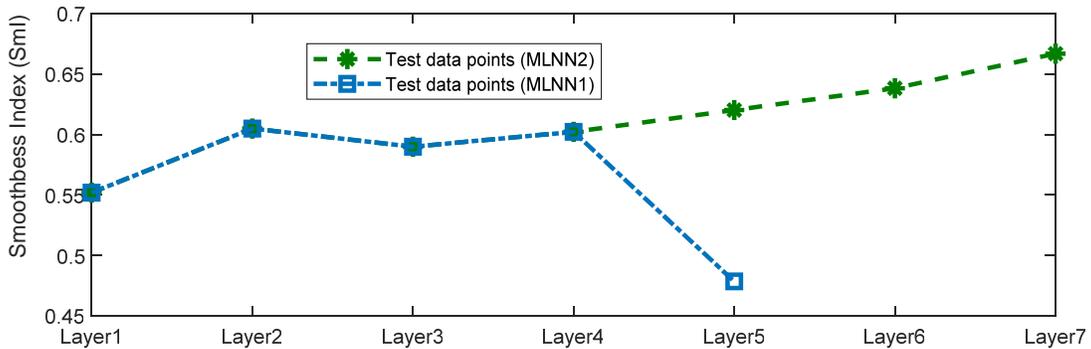

**Figure 16:** Evaluation of test dataflow through layers of two MLNNs in the prediction of Boston housing price

Figures 15 and 16 shows the plots SmI through layers of MLNN1 and MLNN2 respectively for training and test data. From Figures 15 and 16, the following details are revealed:

(1) The normalization and encoding layers at both networks have improved the quality of data by increasing SmI.
(2) In MLNN1 the linear layer (layer 5) causes a considerable drop on the SmI and the quality of dataflow decreases.
(3) In contrast, in MLNN2, the designed fully connected layers provide a considerable rise on SmI and the quality of dataflow increases.
(4) As a result, MLNN2 provides better dataflow quality for "Boston housing price" and the given RMSEs in Table 6 confirm it.
(5) If some other effective inputs were accessible for this perdition problem, the smoothness indices could be more than they were seen in Figures 14 and 15.

## 5. Conclusion

Two straightforward effective indices were proposed to evaluate the input data and the dataflow through layers of deep neural networks(DNNs), in classification and regression problems. For this mean, separation and smoothness concepts were explained and according to them, two numerical indices were proposed respectively for classification and regression problems. The relation between separation and smoothness was explained and to consider more restrict concepts of separation and smoothness, extended indices were introduced. Some applications of using two proposed indices were explained through four case studies. In the first case study, it was shown that the proposed indices could sort input data from more challenging to less challenging. In the first example for some known classification datasets, it was shown that "Cifar 100"," Cifar 10", "MNIST-Fashion" and "MNIST-Digits" are respectively sorted from more challenging to less challenging. Similarly, in the second example, it was shown that "Daily Sunspot number ", "Monthly Sunspot number ", "Electrical loads", "Gold price" and "Oil Price" are sorted respectively from more to less challenging time series for prediction purposes. In the second case study, through two examples: classification of "Cifar10" and "Cifar100" by "VGG 16" network, it was shown that the proposed separation index could be used to evaluate and compare each layer of a deep neural network in improving the quality of dataflow. In the third case study, it was shown that the correct classification rate and the separation index are almost equivalent, especially for when the separation index increases. In the last case, the proposed smoothness index was utilized to compare two designed multi-layer neural networks to predict the "Boston housing price". The advantages of using the proposed evaluation indices are not limited only to evaluate and compare the datasets or the performance of the layers in pre-trained networks. They also can be utilized in network compressing and design of deep neural networks in choosing an appropriate number of layers and other hyperparameters.